\documentclass[10pt]{article}
\usepackage[hidelinks]{hyperref}
\usepackage{ijcai19}

\usepackage[linesnumbered,ruled]{algorithm2e}
\makeatletter
\newcommand{\removelatexerror}{\let\@latex@error\@gobble}
\makeatother

\usepackage{lipsum}
\usepackage{booktabs} 

\usepackage{fouriernc} 
\usepackage{stfloats}
\usepackage[skip=2pt]{caption}
\usepackage{times}
\usepackage{soul}
\usepackage{url}

\usepackage[utf8]{inputenc}
\usepackage{caption}
\usepackage{graphicx}

\usepackage[dvipsnames]{xcolor}

\usepackage{amsmath,amssymb}

\title{Closed-Loop Memory GAN for Continual Learning\footnote{Proceedings of the 28th International Joint Conference on
Artificial Intelligence (IJCAI 2019).}}

\author{
Amanda Rios \textnormal{and} \ Laurent Itti\\
\affiliations
University of Southern California, Los Angeles, USA\\
\emails
\{amandari, itti\}@usc.edu
}

\begin{document}
\maketitle

\begin{abstract}
Sequential learning of tasks using gradient descent leads to an unremitting decline in the accuracy of tasks for which training data is no longer available, termed catastrophic forgetting. Generative models have been explored as a means to approximate the distribution of old tasks and bypass storage of real data. Here we propose a cumulative closed-loop memory replay GAN (CloGAN) provided with external regularization by a small memory unit selected for maximum sample diversity. We evaluate incremental class learning using a notoriously hard paradigm, “single-headed learning,” in which each task is a disjoint subset of classes in the overall dataset, and performance is evaluated on all previous classes. First, we show that when constructing a dynamic memory unit to preserve sample heterogeneity, model performance asymptotically approaches training on the full dataset. We then show that using a stochastic generator to continuously output fresh new images during training increases performance significantly further meanwhile generating quality images. We compare our approach to several baselines including fine-tuning by gradient descent (FGD), Elastic Weight Consolidation (EWC), Deep Generative Replay (DGR) and Memory Replay GAN(MeRGAN). Our method has very low long-term memory cost, the memory unit, as well as negligible intermediate memory storage.

\end{abstract}

\section{Introduction}

Since early development and throughout life humans are constantly faced with unknowns in the environment which demand a persistent adaptation and expansion of past knowledge. In addition, as knowledge is expanded, learning is often facilitated since objects and tasks are often closely related and interconnected. For instance, during development, infants learn to categorize animals according to dimensions such as size, texture, shape, sound, among others. However, subsequent addition of new species rarely corrupts classification performance on the already learned categories. In fact, learning broad-species domains can aid in finer species discriminatory capability~\cite{flesch2018comparing}.
\par Nonetheless, recreating human-like lifelong continual learning remains a central challenge in Artificial Intelligence. State of the art deep neural networks (DNN) trained to perform supervised continual learning are known to undergo a phenomenon termed “catastrophic forgetting”, which describes a sharp decline in the performance of the model on previously learned tasks as soon as a new task is introduced~\cite{French1999CatastrophicNetworks,mccloskey1989catastrophic,robins1995catastrophic}. This behavior does not come as a surprise if one recalls that in DNNs, learning an input output mapping implies parameterizing the network with an optimal weight set, through loss minimization. Thus, if training data is unavailable for previous tasks, there will be no more loss term for the old data and a weight parametrization may blatantly deviate from the previous optimal state incurring severe memory erasure. 

\section{Prior Work} 
\par In the recent literature, several methods have been proposed aiming to ameliorate catastrophic forgetting. They can be roughly subdivided into 3 groups:  regularization, network-growing and replay approaches. With regularization methods, one constrains the change of learnable parameters to prevent "overwriting" what was previously encoded. For instance,~\cite{Li2017LearningForgetting} perform distillation between multiple realizations of a network at distinct time-points, ensuring that the new weights do not shift significantly from the old. In a similar vein,~\cite{Kirkpatrick2015OvercomingNetworks} operate within a single network model and use a Fisher information matrix computed with saved samples drawn from past tasks, which then acts as a regularizer preserving highly correlated weights. Similarly,~\cite{Zenke2017ContinualIntelligence} use path integrals of loss-derivatives to constrain weights crucial to past tasks, yielding an intermediate parameterization with minimal combined loss. 
\par Alternatively, in region-growing algorithms, the architecture itself is altered to accommodate new tasks followed by retraining. For instance,~\cite{FernandoPathNetNetworks} freeze the most important paths in the network, therefore forcefully preventing forgetting, and incrementally add new network chunks to incorporate new tasks. Lastly, In replay methods, the models no longer preserve a key pathway or weights. In these algorithms, one estimates the distribution of the old data either by saving a small fraction of the original dataset into a memory buffer or by training a generator to mimic the lost data and labels. At each new task, these methods learn by presenting a network with both new images as well as replay of estimated or buffered old images, reverting the continual framework into a multi-task setting and thus alleviating forgetting~\cite{furlanello2016active}. Other works have built on the idea of using a buffer of real data to approximate the past distribution~\cite{Rebuffi2001ICaRLLearning,Lopez-paz2017GradientLearning,nguyen2018variational}.  
\par Yet, despite a growing number of appealing solutions, catastrophic forgetting is not a solved issue. Regularization methods have been shown to perform poorly in single-headed incremental class learning, for instance~\cite{Kemker2018FEARNING,parisi2019continual}, and here we reproduce this limitation in our own results for elastic weight consolidation~\cite{Kirkpatrick2015OvercomingNetworks}. On the other hand, region growing approaches, while usually providing a clean solution for constrained incremental problems, can quickly become memory expensive since they require both an architectural expansion and the storage of at least a portion of old data for retraining. 
\par Likewise, replay methods also run into scalability issues. So far, generative replay models learn a data distribution by resorting to intermediate copy states of the generator. In Deep Generative Replay (DGR) an unconditional GAN is trained at each task to cumulatively generate and discriminate images. Since the proposed GAN is unconditional, they employ an additional classifier (Solver) which is trained in parallel to classify the generated images and assign corresponding labels~\cite{LeeContinualReplay}. During each task switch, DGR makes a copy of the generator and classifier networks and uses them to generate sample images and labels for the old tasks. In Memory Replay GAN (MeRGAN) with joint replay,~\cite{wu2018memory} propose a modification in the DGR framework by substituting the unconditional GAN for an ACGAN, thereby eliminating the need for the additional solver. Copy operations are both expensive and often lead to image quality being degraded through consecutive tasks. Moreover, replicating network states successively is not a fully desirable solution since, from the biological perspective, a human brain cannot produce an “intermediate copy” of itself to transfer knowledge. Lastly, methods which rely rather on small subsets of past data, memory buffers, have shown to yield good results but they do not make explicit how much of the performance is due to the algorithm developed and how much is intrinsically due to the variability included in the buffer unit. 

\section{Closed Loop memory GAN}

\subsection{Model Overview}
In this paper, we propose a hybrid approach between memory buffers and deep generative models aiming to specifically reduce memory costs and maximize both the classification performance and generated image quality throughout training. In our model, there is only one generator and embedded classifier trained cumulatively, with no intermediate copy step. In this framework, as a new task is learned, the old data is approximated by continuously sampling from the generator at its present state, forming a closed loop training paradigm. Of course, since a new task also modifies the parameterization of the generator, this procedure cannot be applied without some verification that the generated images are reasonable approximations of the old distribution that has been lost. Our method tackles this issue by, first, using an image filtering step in which either the classifier or the discriminator is used to assess the sample image quality and, as a result, blocking bad images from entering the training loop. Second, we employ external regularization by constructing a small dynamic memory buffer with real data samples chosen to maximize image heterogeneity and to enforce smoothness in the representation of old classes. The image buffer has fixed memory allotment. Therefore, it is not allowed to grow which requires eliminating some old images to make room for new ones. The sampling for the old data is then always a combination of buffer samples and “on-the-fly” generated samples, which provide a stochastic up-sampling of the memory unit. 

\begin{figure}[t]
    \begin{center}
        \includegraphics[width=8.2cm]{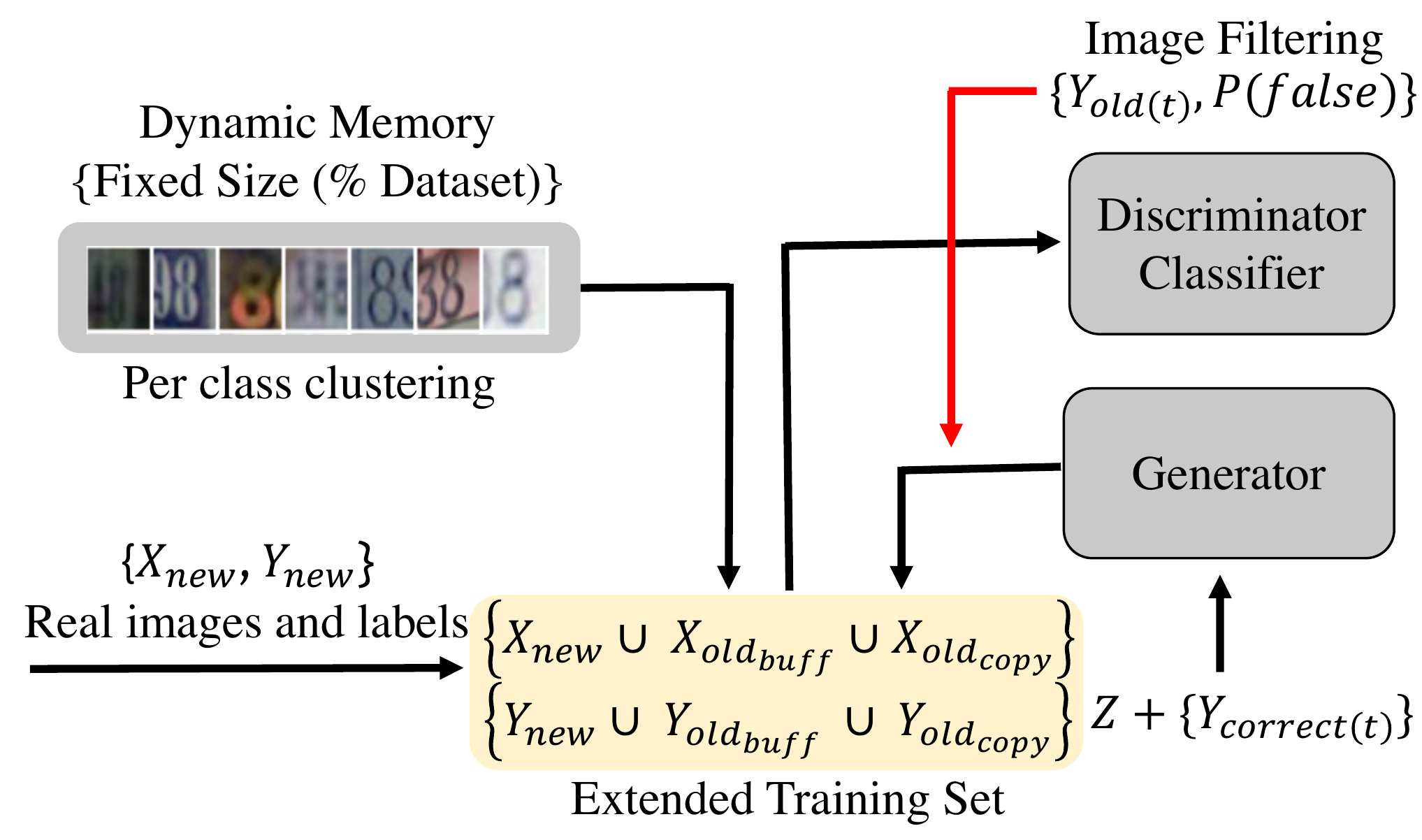}
		\caption{CloGAN Model used for cumulative and continual learning. Past data is sampled from the generator and filtered by the embedded classifier. Old data is a combination of the fresh stochastic generator output and a small memory buffer used to "smoothen" the old data distribution for quality output.}
        \label{fig:model}
    \end{center}
\end{figure}

\subsection{Model Architecture}

A vanilla GAN consists of two networks, a Generator and a Discriminator, competing with each other in a zero-sum game framework. The core block of our model (CloGAN), see figure (1), is a modified GAN termed Auxiliary Conditional Generative Adversarial (AC-GAN)~\cite{Odena2017CYNTHESIS}. The AC-GAN is also composed of 2 networks, but it includes a classifier combined in the same architecture as the discriminator, via an expansion to K+1 output nodes, for K classes plus the original vanilla Real/Fake discriminator output. 
\par In an AC-GAN framework the generator is fed a uniform noise $z \sim p_{z}$ appended with a corresponding class label $c \sim p_c$. Thus, the conditional generator, described by $\theta_{G}$, generates an image $x = G_{\theta_{G}}(z,c)$ and the AC-GAN learns a mapping in which the noise $z$ is independent of the class $c$, enabling multiple class outputs for a fixed noise input. While the generator $\theta_{G}$ is trained to generate images as closely resembling the input image distribution, the discriminator, $\theta_{D}$, is conversely trained to discriminate these generated images as fake, loss $L_{FT}$. The embedded classifier, $\theta_{C}$, shares most weights with the discriminator and generates a label prediction which, if incorrect, contributes to the overall loss of both generator and discriminator, ${L_{C}}$.  Overall, an AC-GAN is easier to train than a conventional vanilla GAN while also producing higher quality images. The loss functions are given as follows in (1) and (4) for generator and discriminator/classifier respectively.
\begin{align}
\theta_{G}^{*} = \min_{\theta_{G}}(L_{FT}^{G}(\theta, X) + L_{C}^{G}(\theta, X))\\
L_{FT}^{G}(\theta, X) = - \mathbb{E}_{z\sim p_{z}, c\sim p_{c}}[D_{\theta_{D}}(G_{\theta_{G}}(z,c))]\\
L_{C}^{G}(\theta, X) = - \mathbb{E}_{z\sim p_{z}, c\sim p_{c}}[y_{c}log(C_{\theta_{C}}(G_{\theta_{G}}(z,c)))]
\end{align}
\begin{align}
\theta_{D}^{*}, \theta_{C}^{*}= \min_{\theta_{D}, \theta{C}}(L_{FT}^{D}(\theta, X) + L_{C}^{D}(\theta, X))\\
L_{FT}^{D}(\theta, X) = \mathbb{E}_{z\sim p_{z}, c\sim p_{c}}[D_{\theta_{D}}(G_{\theta_{G}}(z,c))] -  \\ \mathbb{E}_{(x,c)\sim X}[D_{\theta_{D}}(x)]\\
L_{C}^{D}(\theta, X) = - \mathbb{E}_{(x,c)\sim X}[C_{\theta_{C}}(G_{\theta_{G}}(z,c))]
\end{align}

Note that a plausible alternative to using a GAN would be to use a variational auto encoder (VAE) instead~\cite{kingma2014semi}. However, in our testing, we have not been able to achieve results with a VAE as good as those presented here using a GAN. Hence, in the following, we restrict our analysis to approaches based on GAN. Details of the implementation can be found in the \href{https://drive.google.com/open?id=1aRYkHq5GdX5cDlFbs7coYb9KyTWbHine}{supplementary materials link}~\cite{supplementary}. 

\subsection{Closed-Loop Training with Replay}

In the continual learning setting, our method approximates the likelihood of old data by employing CloGAN to continuously output fresh new images at each mini-batch during training. A combination of image filtering and external regularization by an image memory buffer confer stability to the closed-loop procedure. At each task, our model is trained using an extended dataset which includes real images for the new task, GAN replayed images for old tasks, and memory images, forming an extended training set $S_t$ (8), see figure 2. The memory component can be given a weighted importance, $\lambda_{mem}$. The network is then trained by minimizing (9,10).

\begin{align}
S_t = {S_{t}^{real} \cup S_{t-1}^{GAN} \cup \lambda_{mem} S_{t-1}^{memory}}\\
\min_{\theta_{D}^t,\theta_{C}^t}(L_{R}^{D}(\theta, S_t) + L_{C}^{D}(\theta, S_t))\\
\min_{\theta_{G}^t}(L_{R}^{G}(\theta, S_t) + L_{C}^{G}(\theta, S_t))
\end{align}

\subsection{Image Filtering}

At each mini-batch, the generator outputs fresh images approximating samples from old tasks, with the intent of producing a stochastic up-sampling of the reduced memory core. However, since these images are then used as training data in a closed loop, they have to be of the best quality possible to minimize error propagation. Thus, at each generation step, images are assessed for their quality and "filtered" out if they do not correspond to the standard. 
\par Here, we use the embedded classifier in CloGAN to generate a prediction for the conditional image. If this prediction does not match the conditioning label, the image is filtered out. When old images are generated for closed-loop replay, they are sampled from a model which has already previously converged for generation and classification of old tasks. The rationale behind this evaluation is that images which are missclassified have a higher probability of being distorted because of the ongoing training of the new task, and of deviating too grossly from the original distribution. We term this method Class-Conditioned Filtering (CFM).

\par In addition to CFM, we implemented a more complex procedure, "Discriminator Rejection Sampling" (DRS) proposed in~\cite{Azadi2018DiscriminatorSampling}. The latter employs the discriminator of an AC-GAN to approximately correct errors in the GAN generated distribution. Details of the implementation can be found in the \href{https://drive.google.com/open?id=1aRYkHq5GdX5cDlFbs7coYb9KyTWbHine}{supplementary materials link}. We compare both to a baseline case for DRS which rejects a sample if the output from its discriminator logit layer has a score below some threshold, Soft rejection Filtering (SRF)~\cite{mackay2003information}. Overall, we found that CFM, DRS and SFR perform equivalently well. A table with comparisons is included in the \href{https://drive.google.com/open?id=1aRYkHq5GdX5cDlFbs7coYb9KyTWbHine}{supplementary materials link}. Hence, since CFM has a much faster running time, we opted for carrying out only class conditional filtering in our final model. 
    
    
\begin{figure}[t]
\removelatexerror
    \begin{algorithm}[H]
        \SetKwInOut{Input}{Input}
        \SetKwInOut{Require}{Require}
        \Input{
        Data $S_t^{real}, ..., S_T^{real}$\;}
        \Require{
        $T$: Number of Tasks\;
        $I_t$ : Number of iterations\;
        $B$ : Buffer Size\; 
        $K_c$ : Number of clusters per class\;
        $\lambda_{mem}$ : Memory importance\; 
        }
        $\theta_{G}^{*},\theta_{D,C}^{*},\theta_{C}^{*} \leftarrow$ \text{Train AC-GAN($S_{t=1}$) for $i=1$ to $I_1$}\\
        $S_{t=1}^{memory} \leftarrow$ \text{BufferConstruct $(K, B,S_{t}^{real})$}\\
    
        \For{$t \leftarrow 2$ \KwTo $T$}{
        $S_t^{*} = S_{t}^{real} \cup \lambda_{mem}S_{t-1}^{memory}$\\
        \For{$i \leftarrow 1$ \KwTo $I_t$}{
        $S_{t}^{i*} \leftarrow$ \text{Batch}$(S_t^{*})$\\
    
        $S_{t-1}^{GAN} \leftarrow$ \text{Forward $(G(z,y^c_{t-1}))$}\\
        $S_{t-1}^{GAN} \leftarrow$ \text{Filter ($S_{t-1}^{GAN}$)}\\
        $S_{t}^{i} \leftarrow S_{t}^{i*} \cup S_{t-1}^{GAN}$\\
        $\theta_{G}^{*},\theta_{D,C}^{*},\theta_{C}^{*} \leftarrow$ \text{Train AC-GAN($S_{t}^i$)}\\
        }
        $S_{t}^{memory} \leftarrow$ \text{BufferConstruct $(K, B, S_{t-1}^{memory},S_{t}^{real})$}\\}
    \caption{CloGAN Train}
    \end{algorithm}
\caption{Training Algorithm. Procedure $Train$ is described in section 3.3; $Filter$ in 3.4; $BufferConstruct$ in 3.5}
\end{figure}
\subsection{Dynamic Memory Buffer}
We fill a small memory buffer with samples and labels of original past data to perform external regularization. The memory can be seen as a stable reference frame throughout training that enforces a "smoothness" in the representation for each class. At each task, a selection method is employed to choose the samples from the new task which will go into the buffer, with the aim to maximize sample heterogeneity. Also, since a buffer has fixed size, this selection method is further used to determine which of the old task samples will be removed to make space for the incoming new data, employing again the heuristics of sample heterogeneity. Several buffer selection strategies were initially experimented but the best selection scheme was K-means clustering per class, both at image insertion and removal. In more details, the construction scheme is as follows: at the end of each current task, a k-centers algorithm is run per each class in the current tasks's training labels, super-labeling each image as one of K clusters. At the time of insertion into the memory buffer, we select equal numbers of image samples from each class-specific cluster. Additionally, if the buffer is full we compute the space needed for new images and remove an equivalent number of old images. We do this by assessing their stored super-cluster labels and removing equal amounts of samples per cluster, thereby preserving heterogeneity. By storing the per-class, cluster assignment superlabels we also avoid repeating the clustering operation. 

\subsection{Continual Learning Baselines}

We evaluate other continual learning algorithms as baseline comparisons. We implement Elastic Weight Consolidation (EWC; Kirkpatrick~\shortcite{Kirkpatrick2015OvercomingNetworks}), Deep Generative Replay (DGR; Shin~\shortcite{LeeContinualReplay}) and Memory Replay GAN (MeRGAN; Wu~\shortcite{wu2018memory}) . With DGR, to make our implementation a fair comparison, we use an unconditional GAN with the same architecture and complexity as our CloGAN, except that it has only one Real/Fake output node. For both EWC and DGR, we use a classifier with identical architecture as our embedded classifier/discriminator, but with one fewer output node since a pure classifier does not evaluate Real/Fake attribution. Finally, for MeRGAN we implement an AC-GAN with identical architecture as our CloGAN. 

\section{Experiments}
\subsection{Buffer Selection}
\par We experimented with several buffer selection schemes but they under-performed class-specific K-centers. In the other selection methods, we extracted the logit or softmax layer of the discriminator/classifier network and computed measures such as Kurtosis and Peak-Difference to assess sample heterogeneity. The latter measure corresponds to the difference between softmax scores of the most probable and second most probable class for a given image. As such, we ranked the images according to each measure and kept the images with a probability proportional to their score. In other words, we performed a roulette weighting procedure such as in genetic selection~\cite{goldberg1991comparative}. Table 1 contains performance metrics for 3 buffer selection schemes and no selection (none) during CloGAN incremental class learning using the FASHION dataset with memory buffer of size 0.16\%. 

\begin{table}[h!]
\label{sample-table}
\centering
\scalebox{0.9}{
\begin{tabular}{cc}
\toprule
Method & CloGAN\\
\midrule
Class-Kcenter & \textcolor{red}{75.87 +/- 0.43}\\
Kurtosis  & 64.52 +/- 0.73\\
Peak Difference & 57.74  +/- 0.61 \\
None & 71.03  +/- 1.4\\
\bottomrule
\end{tabular}}
\caption{\% Correct As A Function Of Buffer Selection}
\end{table}

\subsection{Incremental Learning}

We evaluate continual learning as accumulating knowledge of a growing number of disjoint classes, termed incremental learning. Furthermore, we make use of a challenging variation of incremental learning, “single headed learning”. Here, each task is a disjoint subset of classes from the overall dataset. Performance is evaluated for all previous classes, resulting in a 1/K chance level, where K is the number of classes accumulated to that point. We evaluate incremental class learning in 4 datasets: MNIST~\cite{lecun1998gradient}, FASHION~\cite{xiao2017fashion}, SVHN~\cite{netzer2011reading} and E-MNIST~\cite{cohen2017emnist}. The first 3 were subdivided in disjoint subsets of 2 classes per task, with a total of 5 tasks to cover all the label types. E-MNIST, a larger dataset, was divided into tasks of 3 classes, covering 24 different classes in 8 consecutive tasks. To account for the growing number of classes, we create extra output nodes which are incrementally used, which allows us a single head for all tasks. 
\par We distinguish our procedure from Multi-Headed learning~\cite{FarquharTowards2018} in which prediction is constrained to classes in each task. For instance, a multi-headed version of our MNIST test would use and re-use only two output nodes. After training on full disjoint MNIST with 5 tasks of 2 classes each, when evaluating the first task (digits 0 and 1), a multi-headed would only have to decide between digit 0 vs 1, as opposed to a one in ten decision for single-headed. This typically leads to much higher accuracies partially because an output node never becomes completely disabled, as it is always used for the last task. Finally note that a multi-headed network with only 2 output nodes provides an output that needs to be further disambiguated by knowing the task. 

\begin{figure*}[t]
    \begin{center}
        \includegraphics[width=17.5cm]{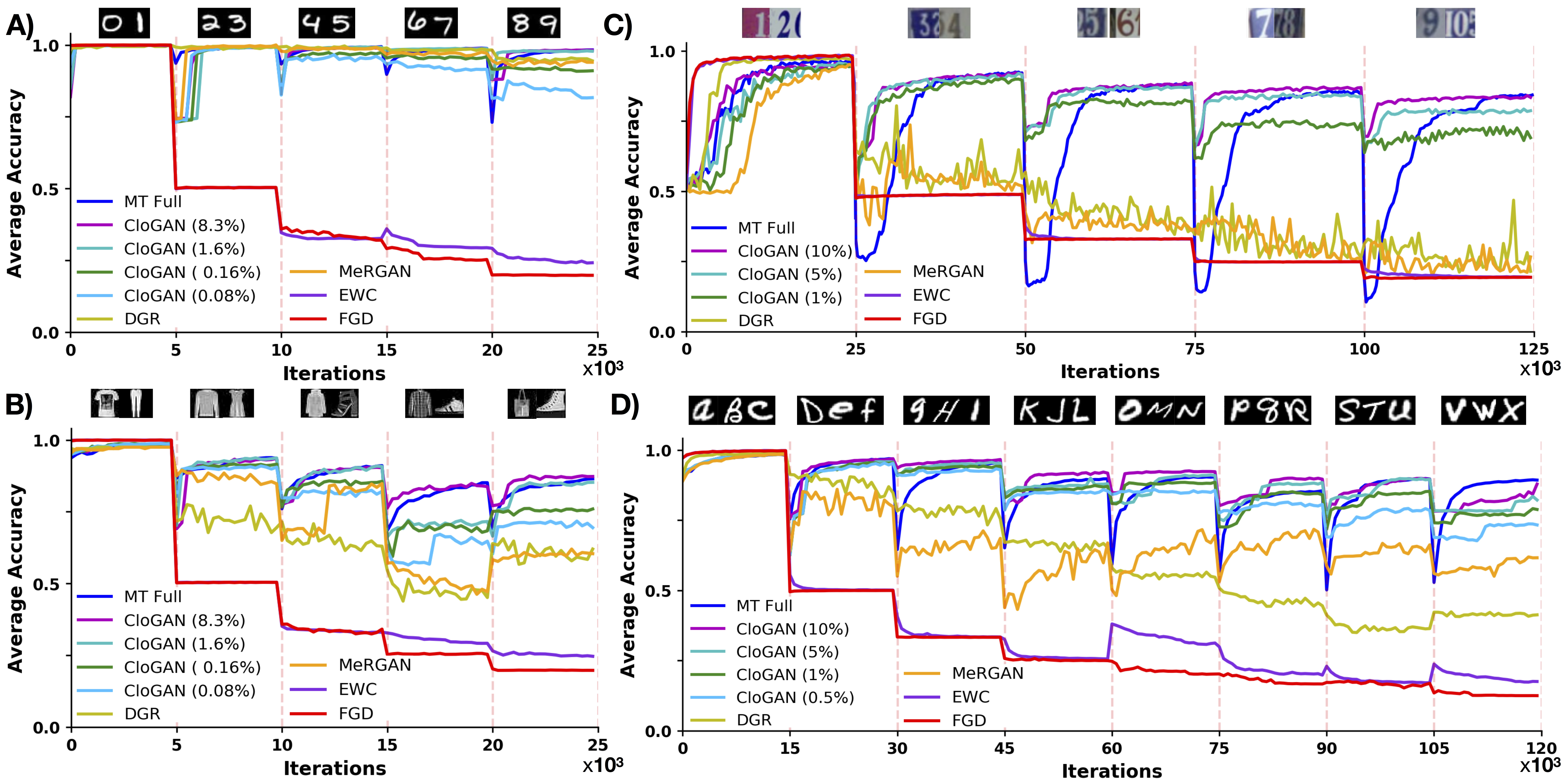}
		\caption{Maximum Average Accuracies across continually learned tasks. A) MNIST; B) Fashion; C) SVHN; D) E-MNIST. The dashed lines indicate start of a new task represented by a disjoint set of classes. We illustrate the performance CloGAN as memory sizes are varied (memory allotment is in parenthesis). In red we show catastrophic forgetting when fine-tuning by gradient descent. In salmon we show multi-task  training until convergence with the full datasets starting from scratch at every task switch (MT-Full). We also show results for EWC, DGR and MeRGAN. }
        \label{fig:model}
    \end{center}
\end{figure*}

\subsubsection{Average Continual Performance}
Figure 3 displays the average performance of CloGAN when varying memory buffer size. Our method avoids catastrophic forgetting even with very small buffer sizes such as 0.08\% (50 images) and 0.16\% (100 images), for both MNIST and FASHION. For the more challenging E-MNIST and SVHN, buffer requirement becomes more demanding. Nonetheless, we obtain superior performance over the competing methods with still very reduced memory sizes: only 0.5\% (576 images) and 1\% (492 images).  
\par Table 2 compares maximum average accuracies after training all tasks, for all methods tested. First, when no memory or GAN sampling is performed  catastrophic forgetting occurs, as exemplified by the \textit{FGD} condition which contains only fine-tuning with gradient descent. Second, EWC accuracy rapidly declines, asymptotically reaching the catastrophic forgetting curve. EWC has already been shown to behave poorly in incremental single-headed paradigms~\cite{Kemker2018FEARNING,FarquharTowards2018}. To further confirm that this degradation of performance was not particular to our implementation, we replicated the permuted-MNIST experiment proposed in the original EWC paper ; and verified that in this learning paradigm EWC performs very well. This discrepancy between the experiments is likely due to the difference in output mapping, see \href{https://drive.google.com/open?id=1aRYkHq5GdX5cDlFbs7coYb9KyTWbHine}{supplementary}.

\begin{table}[h!]
\centering 
\renewcommand{\arraystretch}{1.3} 
\begin{tabular}{p{1.5cm}@{\hspace{0.6cm}}p{1.1cm}p{1.1cm}p{1.1cm}p{1.1cm}} 
\cmidrule(l){1-5} 
\textbf{Method} & Mnist & Fashion & Svhn & Emnist\\ 
\midrule 
MT-full & 98.29 & 86.48  & 84.43 & 89.41\\ 
CloGAN & \textcolor{red}{98.03 (1.6\%)} & \textcolor{red}{85.25 (1.6\%)} & \textcolor{red}{79.30 (5\%)} & \textcolor{red}{83.50 (5\%)}\\ 
CloGAN &  \textcolor{red}{92.26 (0.16\%)}& \textcolor{red}{76.15 (0.16\%)} & \textcolor{red}{73.08 (1\%)} & \textcolor{red}{79.14 (1\%)}\\ 
MeRGAN & 98.25 & 65.62 & 31.94 & 61.92 \\ 
DGR & 94.90 & 62.11 & 46.83 & 42.35 \\ 
EWC & 29.19 & 26.52 & 22.55 & 23.76\\ 
FGD & 19.97 & 20.22 &  19.56 & 14.28 \\ 
\midrule 
\end{tabular}
\caption{Performance after continual learning of all tasks. For CloGAN, memory allotment is in parenthesis} 
\label{tab:template} 
\end{table}

Lastly, we report the accuracies for the deep replay methods, DGR and MeRGAN. For MNIST, both DGR and MeRGAN perform very well, reaching 94.9 \% and 98.25\% whereas CloGAN achieves accuracies of 92.26\% with memory of 0.16\% and 98.03 with (1.6\%). However, for all other datasets, which are significantly harder than MNIST, DGR and MeRGAN both underperform CloGAN by significant amounts. 
\par For SVHN, the most challenging dataset, both DGR and MeRGAN display degraded performance after the first task. This behavior likely has cause in a persistent degradation of generated image quality throughout training. Both methods represent old data exclusively by replayed images from an intermediate generator copy. If the generator cannot produce images which represent the original distribution with high fidelity, the gap in representation capacity can be enlarged and propagated through successive GAN transfer (copy) operations. CloGAN alleviates GAN representation degeneration because it is trained from an extended set containing both replay images from the generator and real images in the buffer. The real images never degenerate and act as an anchor to keep smoothness and quality in the subsequent generated images. DGR has another disadvantage over CloGAN: it does not generate conditioned images, requiring a separate classifier to produce old image labels during training. If that classifier does not have perfect performance, it will inevitably misslabel some images, contributing to error propagation. 

\subsubsection{Stochastic Up-Sampling}
We confirm that CloGAN performs an upsampling of the memory buffer selection by comparing our method to two variations in which the AC-GAN is trained only from a memory buffer, both in continual (Frozen-CloGAN) and multi-task settings (MT). For the latter two conditions there is no closed-loop replay of GAN samples. Furthermore, in the MT setting we re-start training at each task switch.  The results reported in figure 4 correspond to the maximum accuracies achieved for each task for all 3 variations.  We verify that stochastic generation in CloGAN provides an upsampling of the buffer and achieves superior performance to Frozen-CloGAN and MT. We show results for the more challenging datasets, E-MNIST and SVHN. 
\begin{figure*}[t]
    \begin{center}
        \includegraphics[width=17cm]{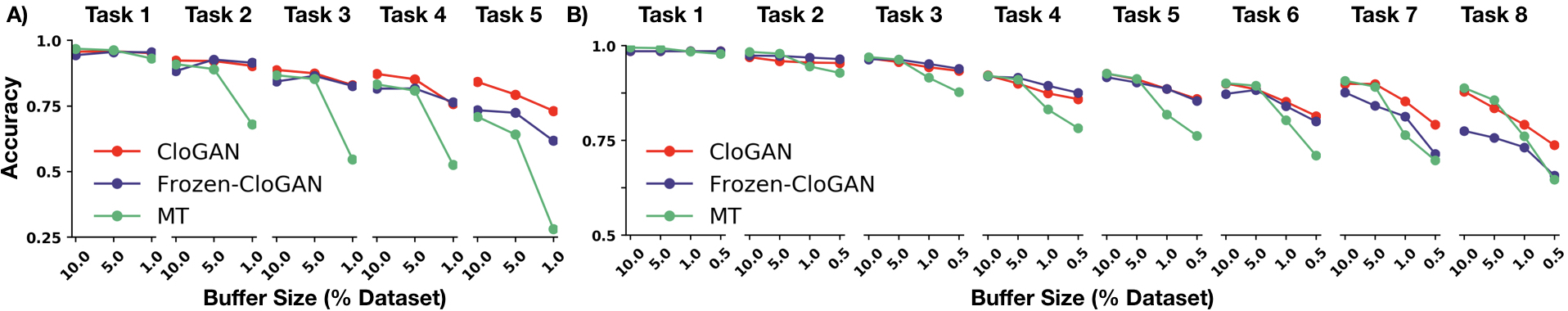}
		\caption{Stochastic Up-Sampling of CloGAN. A) SVHN. B) E-MNIST. The contribution of upsampling is indicated by a positive gap between CloGAN and Frozen-CloGAN as more tasks are learned. We also compare to the MT condition in which training is re-started at each task, eliminating forward-transfer of possible shared task features.}
        \label{fig:model}
    \end{center}
\end{figure*}
\par Upsampling is indicated when a positive gap between CloGAN and Frozen-CloGAN increases as more tasks are added. For SVHN, the last task shows clear gaps between CloGAN Frozen-CloGAN as well as MT (maximum gap of 11.39\% at task 5). Similarly, E-MNIST shows a clear gap in the last two tasks, 7 and 8 (maximum gap of 10.41\% at task 8). Additionally, we show that MT under-performs starting in early tasks due to lack of forward transfer since the networks are re-started from scratch at each task switch. Similar upsampling behavior was observed in MNIST and FASHION, with maximum gaps of 6.84\% and 9.35\% respectively. Additional figures can be found in the \href{https://drive.google.com/open?id=1aRYkHq5GdX5cDlFbs7coYb9KyTWbHine}{supplementary link.}  

\begin{figure}[h!]
    \begin{center}
        \includegraphics[width=8.0cm]{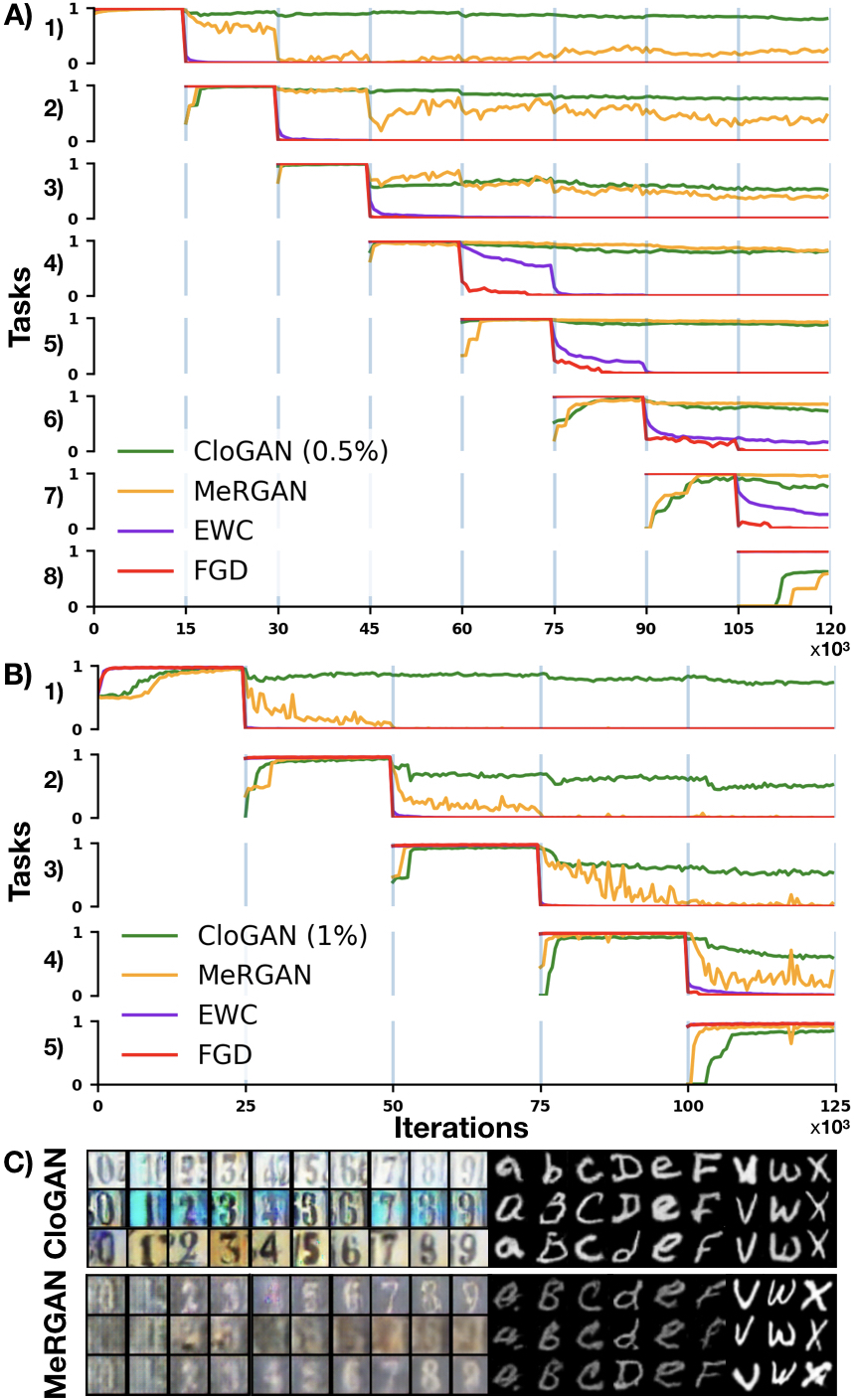}
		\caption{Accuracies per task for EMNIST (A) and SVHN (B). Generated images for SVHN and EMNIST (C). CloGAN preserves performance for early tasks throughout training. MeRGAN has degenerated performance and  image generation for early tasks. In EMNIST, degradation is a darkening in the first tasks (a,...,f) in contrast to the last task (v,w,x).}
        \label{fig:model}
    \end{center}
\end{figure}

\subsubsection{Memory Equivalence of Stochastic Replay}

To further disentangle the contribution of a closed-loop generative replay to the model, we compensate the memory expense of the stochastic generator by allocating all of its memory budget to the episodic buffer. Thus, for each CloGAN we create a NoGenReplay-Equivalent model which equates the generator size (1.6 M float parameters) to images in the episodic memory (2572 if RGB and 6226 if gray). 
\begin{table}[h!]
\centering 
\renewcommand{\arraystretch}{1.3} 
\begin{tabular}{p{1.4cm}@{\hspace{1.0cm}}p{1.0cm}p{1.0cm}p{1.0cm}p{1.0cm}} 
\cmidrule(l){1-5} 
{} & \multicolumn{2}{c}{\textbf{EMNIST}} & \multicolumn{2}{c}{\textbf{SVHN}}\\ 
\textbf{Memory} & 1\% & 5\% & 1\% & 5\%\\ 
\midrule 
CloGAN & \textcolor{red}{79.14} & \textcolor{red}{83.5} & \textcolor{red}{73.08} & \textcolor{red}{79.3}\\ 
NoGenReplay-Equivalent &  77.35 (6.5\%)& 78.68 (10.5\%) & 73.32 (5.2\%) & 73.79 (6.2\%)\\ 
\midrule 
\end{tabular}
\caption{Average performance of CloGAN compared to NoGenReplay-Equivalent, an episodic-only variation. The latter is constructed with a buffer that includes additional images that equate, in bytes, to the size of the stochastic generator in CloGAN.} 
\label{tab:template} 
\end{table}
\par For instance, for CloGAN-1\% trained in EMNIST (gray images) we create the NoGenReplay-Equivalent-6.5\% (buffer of 1\%+6226=6.5\%) and obtain 79.14\% correct for our method versus 77.35\% for the no replay condition. Similarly, for CloGAN-5\% we achieve 83.5\% whereas NoGenReplay-Equivalent-10.5\% yields 78.68\%. Hence, we show that including replay beats just using a larger episodic buffer of equivalent memory. Results are included in table 3.

\subsubsection{Per Task Performance}
In figure 5A,B), we exhibit per task accuracies along time. Here, CloGAN is shown to produce stable performance throughout consecutive tasks. For both E-MNIST and SVHN all past tasks maintain high accuracies consistently throughout learning of new classes. For example in EMNIST, task 1 preserves its accuracy at 84.33\% despite the learning of 7 other tasks in succession. Likewise, SVHN task 1 has an accuracy of 83.87 \%. The results are significantly higher when compared to the baseline of catastrophic forgetting and EWC. Moreover, we also display performance for MeRGAN. In E-MNIST, MeRGAN accuracies for tasks 1 and 2 are clearly underperforming CloGAN at the end of training, likely due to image degradation from GAN to GAN transfer. 
\par In figure 5C), we show generated images by CloGAN and MeRGAN for both SVHN and E-MNIST. We list images taken after training of all tasks. For SVHN we list all classes cumulatively learned. For E-MNIST, since there are 24 classes, we limit the display to the two first tasks as well as the last task (8th). For CloGAN, we find that images are sharp even when using small memory sizes, 1\% - SHVN and 0.5\% - EMNIST. This is true for beginning tasks as well as latter tasks. In contrast, in MeRGAN former taks are sharply more degenerated than latter ones. In EMNIST this can be seen by an overall darkening of letters $a$ through $f$.


\subsubsection{Copy-CloGAN}
We tested a variant of our model, Copy-CloGAN, in which the generator is copied at each task switch. In Copy-CloGAN the stochastic replay samples come from the frozen copied generator and the remaining replay is from the CloGAN episodic memory buffer. In order to properly evaluate the performance of this new model, we account for the extra memory usage by calculating the size in bytes of an extra generator: 1.6M float parameters (6.4 Mbyte). Accordingly, we create a new equiv-CloGAN with a larger episodic buffer to compensate for the duplicate generator of copy-CloGAN. We use the same calculation as described in the construction of the NonGenReplay-Equivalent variant previously described, adding either 2572 RGB or 6226gray images to equate 6.4 Mbytes of extra memory load. 
\begin{table}[h!]
\centering 
\renewcommand{\arraystretch}{1.3} 
\begin{tabular}{p{1.3cm}@{\hspace{0.9cm}}p{1.0cm}p{1.0cm}p{1.0cm}p{1.0cm}} 
\cmidrule(l){1-5} 
{} & \multicolumn{2}{c}{\textbf{EMNIST}} & \multicolumn{2}{c}{\textbf{SVHN}}\\ 
\textbf{Memory} & 1\% & 5\% & 1\% & 5\%\\ 
\midrule 
CloGAN & 79.14 & 83.5 & 73.08 & 79.3\\ 
copy-CloGAN &  81.88 & 87.20 & 73.19 & 80.44\\ 
equiv-CloGAN &  \textcolor{red}{82.60 (6.5\%)}& \textcolor{red}{87.89 (10.5\%)} & \textcolor{red}{81.01 (5.2\%)} & \textcolor{red}{83.74 (6.2\%)}\\ 
\midrule 
\end{tabular}
\caption{Average performances of copy-CloGAN, CloGAN and CloGAN-equiv for different buffer usages. Copy-CloGAN is identical to CloGAN except it copies the generator at each task switch. CloGAN-equiv contains a larger episodic buffer to compensate for the duplicate generator of copy-CloGAN. Notice that while copy-CloGAN performs better than CloGAN for a buffer of same size, it still underperforms equiv-CloGAN, which is built to have the same memory usage. } 
\label{tab:template} 
\end{table}

\par For a given episodic memory size (e.g., 1\%), we compare CloGAN, copy-CloGAN, and equiv-CloGAN. Overall, the copy operation provided a small increase in performance but only when the buffer sizes were held constant, for instance, when trained with SVHN, CloGAN-1\% achieves accuracy of 73.08\% and copy-CloGAN-1\% of 73.19\%. However, when compensating the extra memory usage via buffer augmentation, Copy-CloGAN underperformed equiv-CloGAN-5.2\%, with the latter yielding a 81.01\% correct performance, the highest between the 3 compared models. Thus, on balance, the copy operation did not surpass our approach.

\section{Conclusion}
In conclusion, we have shown how using very small buffers in conjunction with stochastic replay can give rise to superior performance compared to simple gradient descent, EWC or other replay methods. In our model, CloGAN, the memory buffer acts as an external regularization for the generator, counteracting image degradation through time. Our approach is relatively easy to implement and necessitates only low computation (no full retraining) and memory (small buffer), making it ideal to enable life-long learning on resource-constrained mobile (at the edge) devices.


\section*{Acknowledgments}  
This work was supported by the National Science Foundation (grant number 
CCF-1317433), C-BRIC (one of six centers in JUMP, a Semiconductor 
Research Corporation (SRC) program sponsored by DARPA), and the Intel 
Corporation. The authors affirm that the views expressed herein are 
solely their own, and do not represent the views of the United States 
government or any agency thereof.

\bibliographystyle{named}
\bibliography{ijcai19-multiauthor}



\end{document}